\documentclass{article}

     \usepackage[final]{neurips_2023}

\usepackage[utf8]{inputenc} 
\usepackage[T1]{fontenc}    
\usepackage{hyperref}       
\usepackage{cleveref}
\usepackage{url}            
\usepackage{booktabs}       
\usepackage{amsfonts}       
\usepackage{nicefrac}       
\usepackage{microtype}      
\usepackage{xcolor}         
\usepackage{graphicx}

\title{Manipulating Sparse Double Descent}

\author{%
  Ya Shi~Zhang\thanks{\url{https://yashizhang.github.io/}} \\
  Department of Pure Mathematics and Mathematical Statistics\\
  University of Cambridge\\
  \texttt{ysz23@cam.ac.uk} \\
}

\begin{document}

\maketitle

\begin{abstract}
  This paper investigates the double descent phenomenon in two-layer neural networks, focusing on the role of L1 regularization and representation dimensions. It explores an alternative double descent phenomenon, named 'sparse double descent'. The study emphasizes the complex relationship between model complexity, sparsity, and generalization, and suggests further research into more diverse models and datasets. The findings contribute to a deeper understanding of neural network training and optimization. The code is available at \url{https://github.com/yashizhang/sparsedoubledescent}.
\end{abstract}

\section{Introduction}

\begin{figure}[t!]
  \centering
  \includegraphics[width=0.9\textwidth]{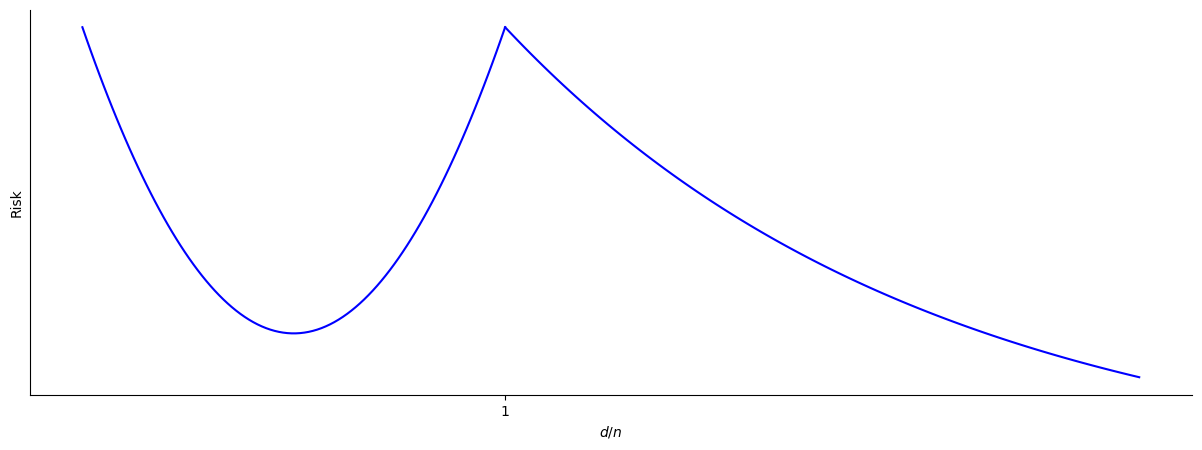}
  \caption{Simple depiction of the double descent phenomenon. $d$ refers to the number of the parameters in the model and $n$ refers to the number of training data points.}
  \label{fig:doubledescent}
\end{figure}

In the modern era of deep learning, it is commonplace for practitioners to specify models that have \emph{many more} parameters than the number of training data. This is due to a statistically non-intuitive phenomenon known as \emph{double descent} \citep{belkin_reconciling_2019}. In a first course on regression, we are often taught the dangers of \emph{overfitting}, when the model fits to the noise in the data and severely deteriorates performance across the entire population. For example, consider polynomial regression, where the degree of the polynomial is significantly higher than the number of training data points. This leads to the notion of a trade-off between the bias and variance of the model, inducing a U-shaped curve (depicted in \cref{fig:doubledescent}) where the vertical axis depicts the population risk and the horizontal axis depicts the ratio between the number of model parameters and number of data points. 

Recently, \citet{curth2023a} has attempted to bridge the gap of our understanding on the prevalence of the double descent phenomenon in non-deep learning models, such as trees, boosting, and linear regression --- methods known as \emph{smoothers}. Notably, analysis of these model classes affirmed the wisdom that model complexity and capacity is not necessarily an affine function of the number of model parameters. Rather, there are higher dimensional substructures in the set of model parameters such that increasing model parameters in certain directions do not cause overfitting. 

In this paper, we examine the possibility of controlling the double descent phenomenon on simple two layer neural networks with varying representation dimension. We view the first layer as a learned approximation of the `ground truth kernel,' and the second layer as a linear classifier. Particularly, we extend \citet{curth2023a} by first considering LASSO regressors, then implementing LASSO via $L_1$ regularization in neural network training as a sparsifier. This is motivated by the perspective of the $L_1$ regularizer as a convex surrogate for the ``$L_0$'' objective, aiming to prune down the number of parameters active in the network. Benefits of pruning and/or sparsification include (1) computational and storage savings, (2) model capacity regularization motivated by Occam's razor \citep{LeCunOptimalBrainDamage}, and (3) potential lottery ticket subspaces \citep{frankle2018the}. 

Furthermore, the prevalence of a `sparse double descent' first noted by \citet{SparseDoubleDescent} --- as model sparsity increases, test performance first decreases then increases (before finally decreasing as sparsity approaches 100\%) --- opens the possibility of model sparsity also exhibiting many intrinsic complexity axes. 

\section{Background}
\label{sec:background}
Consider the supervised learning problem, we are given training data $(x_i,y_i)_{i=1}^n$. In a family of functions $\mathcal{F}:=\{f_\theta(\cdot)\mid\theta\in\mathbb{R}^d\}$ parameterized by $\theta\in\mathbb{R}^d$, we would like to choose a $\theta^*$ such that the function induced $f_{\theta^*}(\cdot)$ minimizes some loss function over the training data, usually of the form $\mathcal{L}(\theta):=\sum_{i=1}^n\ell(f_\theta(x_i),y_i)$. Depending on whether the task is classification or regression, common choices for $\ell(\cdot,\cdot)$ would be the cross entropy loss or quadratic loss, respectively. 

We now modify our loss function by adding a \emph{regularizer}, typically of the form $\alpha||\theta||_p$ for some $p\in(0,\infty]$ and $\alpha\in\mathbb{R}^+$. In this paper we focus exclusively on the case of $p=1$. This $L_1$ penalty --- inspired by the LASSO regressor in linear regression --- inductively biases the network to learn parametrizations that are sparse and well-generalizing. The former is due to the $L_1$ penalty being the projection of the $L_0$ penalty (counting the number of non-zero elements in a vector) to the space of convex functions \citep{l0convexification}. 

If our family of functions are neural networks with $L>1$ layers, then we can view the first $L-1$ layers as a learnable approximation to some unknown, `ground truth' kernel. Through this perspective, we can view the training of neural networks as first learning an appropriate kernel function, then performing linear regression on the kernelized inputs. This is theoretically supported in the lazy training regime \citep{DBLP:conf/iclr/AtanasovBP22, Geiger_2020, jacot1}, and empirically supported via many observed phenomena such as the prevalence of Neural Collapse observed in \citet{Papyan_Han_Donoho_2020}.

\section{Experiments}
\label{sec:experiments}
For our experiments, we will be training a two-layer multilayer perceptron with ReLU non-linearity on the MNIST dataset \citep{lecun2010mnist} using stochastic gradient descent. The empirical experiments comprise of modifying two components of the training procedure to observe a few interesting phenomena. The first is the regularization coefficient $\alpha$. As we increase $\alpha$, we place heavier emphasis on sparsity when training the neural network. The second is the intermediate layers' dimension, which we refer to as the \emph{kernel dimension}. By varying this, we are essentially changing the learned kernel's intrinsic dimension.

\subsection{Results}
\label{sec:results}
As seen in \cref{fig:experiments}, for each kernel dimension, we train separate neural networks with varying $\alpha$. In the top right corner, we have set the middle layer of our two-layer multi-layer perceptron to have $5$ neurons. In the top middle, we have 10 neurons. Following this, we also show results for 25, 50, 75, and 100 intermediate neurons. 

We vary the coefficient of the $L_1$ regularizer to first observe, for all choices of number of intermediate neurons, a version of the sparse double descent phenomenon. 
We also observe the existence of many ascents and descents as we decrease the number of intermediate neurons. 
Finally, we see that the location of the minima is invariant with respect to $\alpha$. This seems to suggest the independence of the sparse double descent phenomena with respect to each layer's width in a fixed neural network architecture. 

\begin{table}[h!]
  \caption{The experiment results as detailed in \cref{sec:experiments}.}
  \label{fig:experiments}
  \centering
  \begin{tabular}{cc}
    \includegraphics[width=0.45\linewidth]{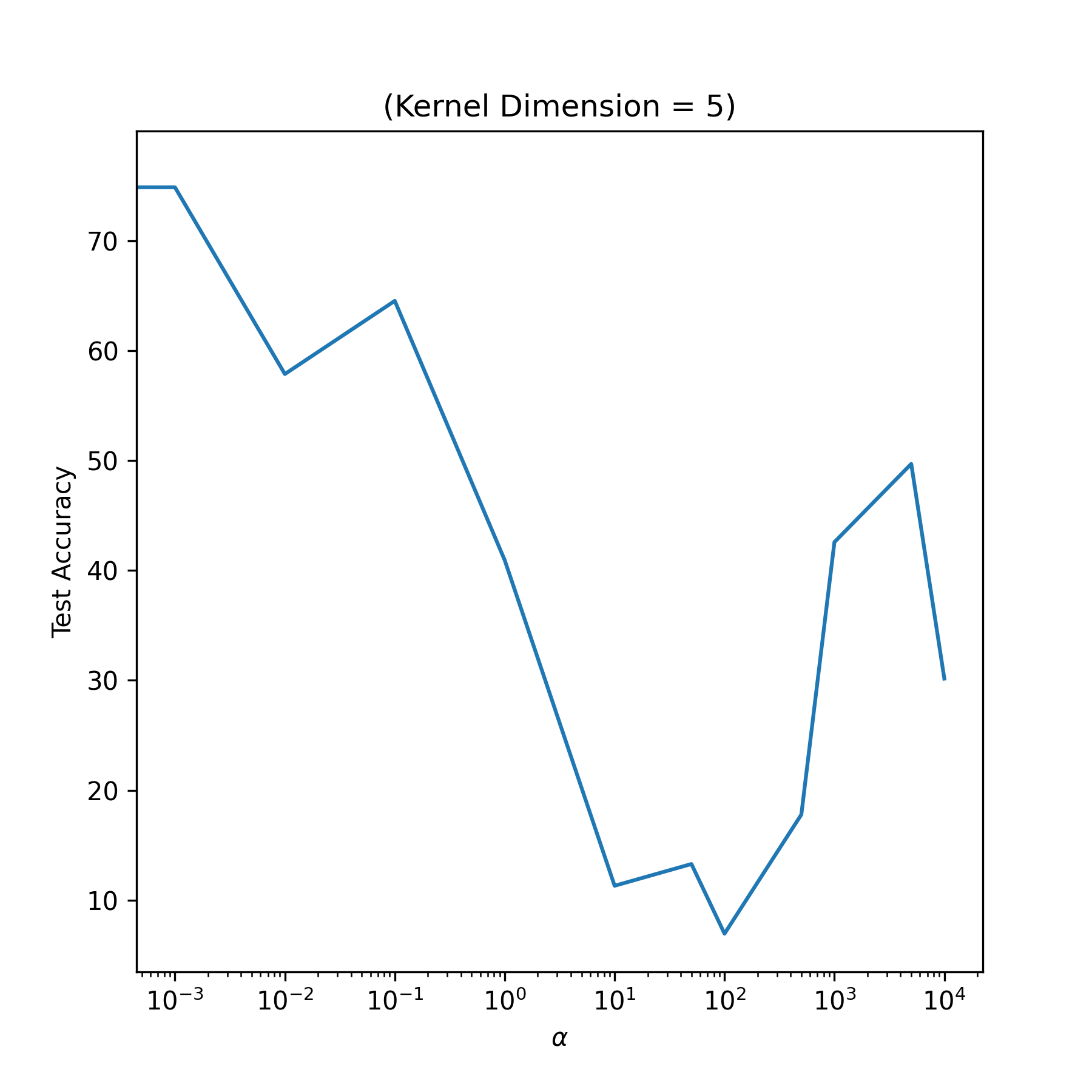}
    & \includegraphics[width=0.45\linewidth]{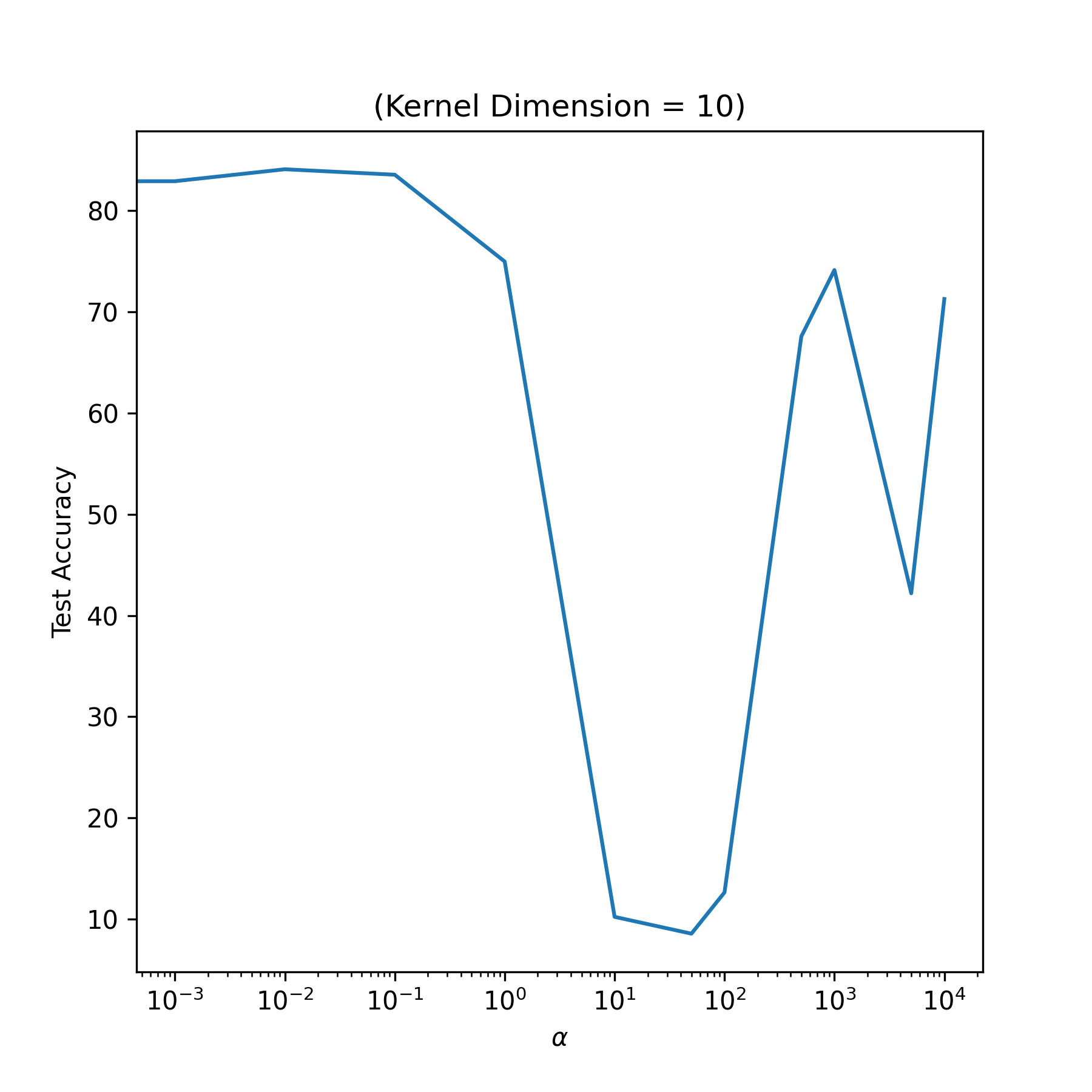} \\
    \includegraphics[width=0.45\linewidth]{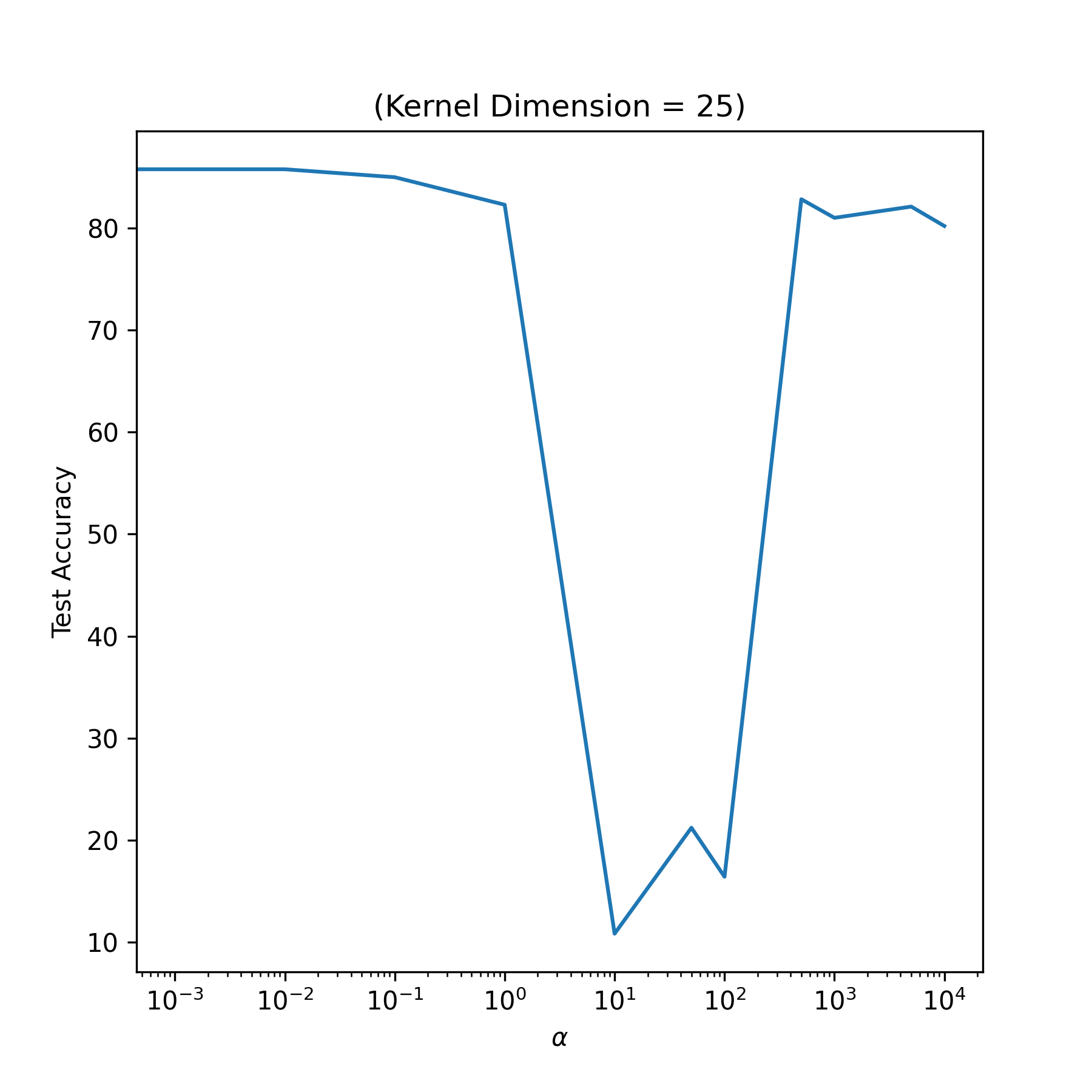} 
    & \includegraphics[width=0.45\linewidth]{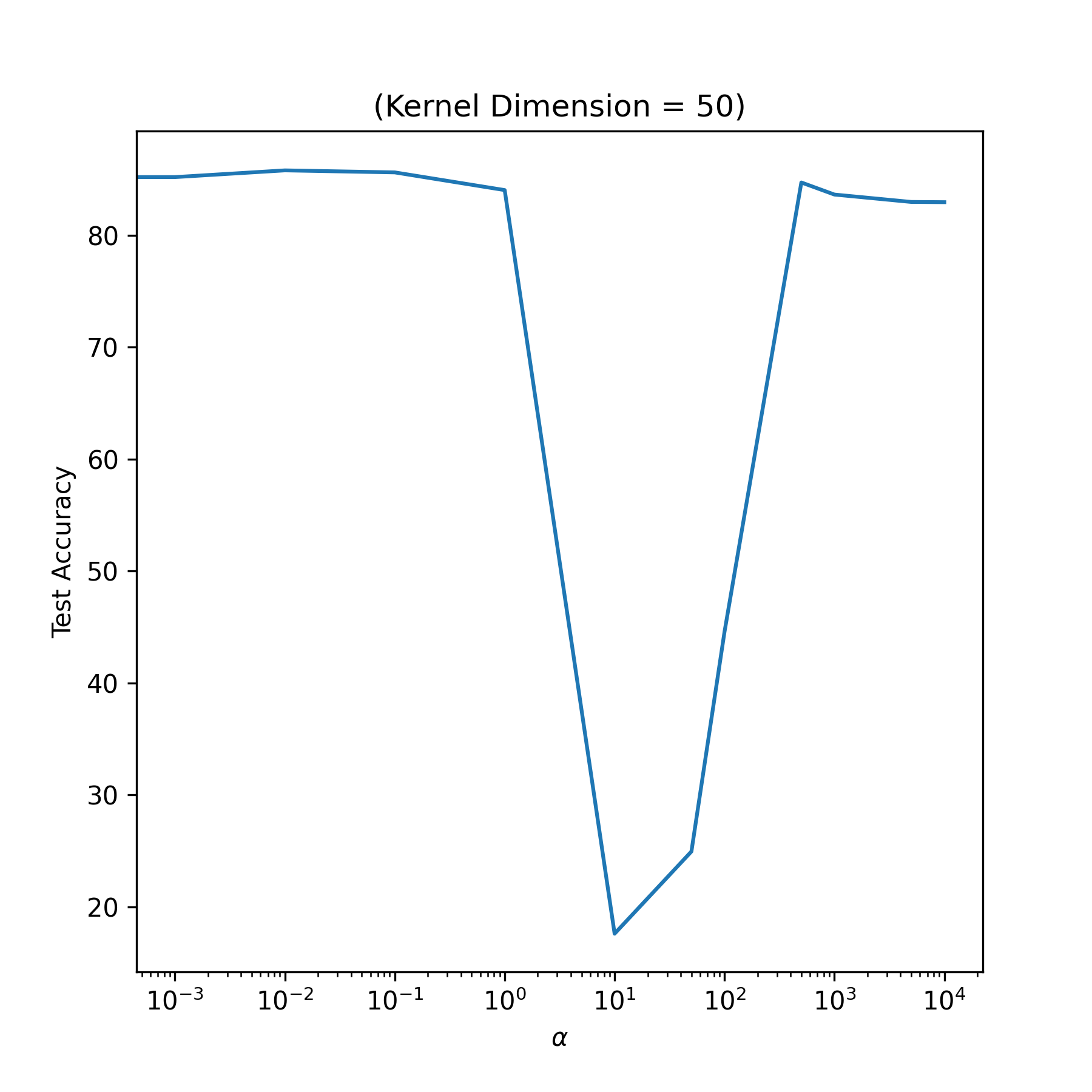} \\
    \includegraphics[width=0.45\linewidth]{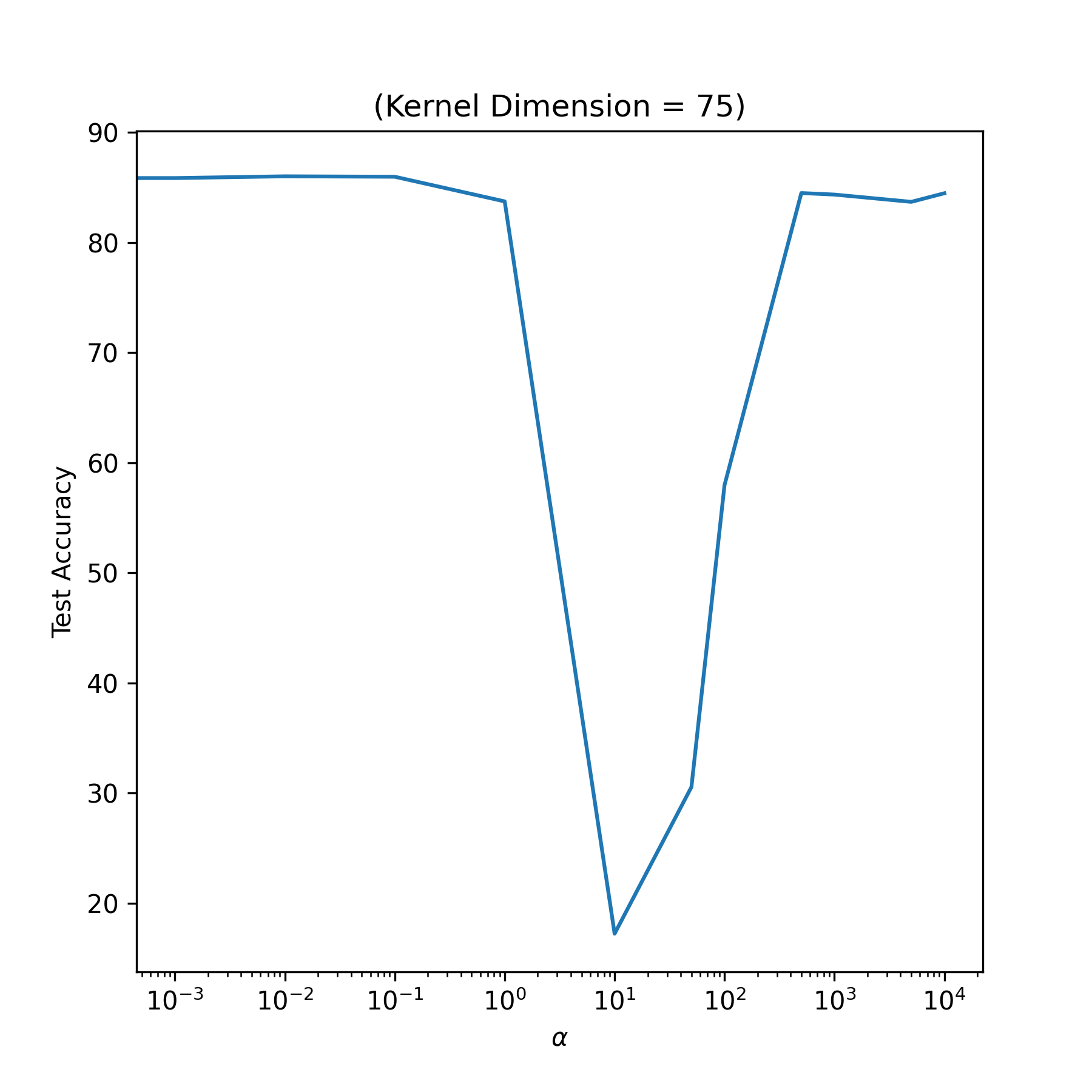} 
    & \includegraphics[width=0.45\linewidth]{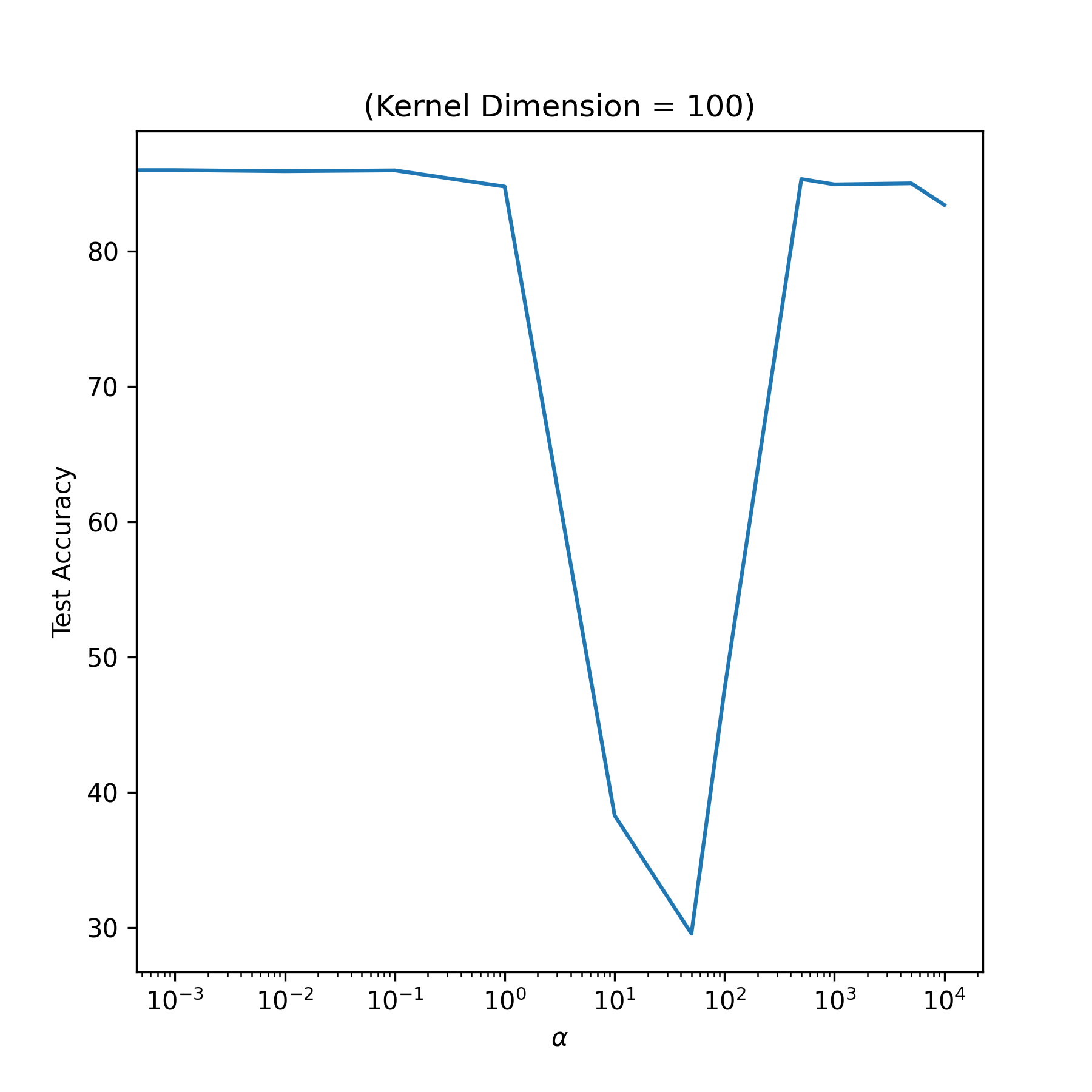} \\
  \end{tabular}
\end{table}

\section{Discussion}
\label{sec:discussion}

Our exploration of the double descent phenomenon in simple two-layer neural networks has yielded some novel insights into model capacity, sparsity, and double descent phenomenona. 
The observed behavior in networks under varying representation dimensions and L1 regularization strenghts underscores a nuanced understanding of the balance between model complexity and overfitting. 

However, this study's focus on specific network architectures and datasets suggests that broader research is required to generalize these findings. Future work should aim to validate these phenomena in more complex models and diverse datasets. Ultimately, this study contributes to the evolving discourse on neural network training, positioning sparse double descent as a key consideration in the pursuit of optimal model performance. This understanding aligns our work with the broader trajectory of machine learning research, highlighting the dynamic and often counter-intuitive nature of model training and generalization.

\subsection{Future Directions}
\label{sec:future}

A future direction for this research could be to expand the experiments to include pretrained vision models fine-tuned on the MNIST dataset. This approach would allow for an exploration of the neural network up to the penultimate layer as a learned kernel, with the retraining of the final layer classifier under different objectives or settings providing insights into various forms of kernel regressions, such as ridge and LASSO.

Additionally, exploring other sparsification and pruning methods could be highly valuable. Theoretical underpinnings for these methods could be provided through Neural Tangent Kernel theory or other kernel theories. Integrating intuitive explanations using concepts like Kolmogorov complexity and Minimum Description Length (MDL) principles would further enhance the understanding. Examining the role of data equivariances and symmetries in these contexts would also contribute to a more comprehensive understanding of the dynamics influencing neural network performance and complexity.

\newpage
\bibliography{references}
\bibliographystyle{plainnat}
\end{document}